\documentclass[10pt,twocolumn,letterpaper]{article}

\usepackage{iccv}
\usepackage{times}
\usepackage{epsfig}
\usepackage{graphicx}
\usepackage{amsmath}
\usepackage{amssymb}
\usepackage{algorithm2e}
\usepackage{verbatim}

\usepackage[pagebackref=true,breaklinks=true,letterpaper=true,colorlinks,bookmarks=false]{hyperref}

\iccvfinalcopy 

\def\iccvPaperID{1347} 

\ificcvfinal\fi
\begin{document}

\title{Generative Adversarial Residual Pairwise Networks for One Shot Learning}

\author{Akshay Mehrotra \hspace{3cm} Ambedkar Dukkipati\\
{\hspace{-1.6cm} \tt\small akshay.mehrotra@csa.iisc.ernet.in \hspace{1.1cm}  \tt\small ad@csa.iisc.ernet.in} \\
Indian Institute of Science\\
Bangalore\\
}

\maketitle

\begin{abstract}
Deep neural networks achieve unprecedented performance levels over many tasks and scale well with large quantities of data, but performance in the low-data regime and tasks like one shot learning still lags behind. While recent work suggests many hypotheses from better optimization to more complicated network structures, in this work we hypothesize that having a learnable and more expressive similarity objective is an essential missing component. Towards overcoming that, we propose a network design inspired by deep residual networks that allows the efficient computation of this more expressive pairwise similarity objective.
Further, we argue that regularization is key in learning with small amounts of data, and propose an additional generator network based on the Generative Adversarial Networks where the discriminator is our residual pairwise network. This provides a strong regularizer by leveraging the generated data samples. The proposed model can generate plausible variations of exemplars over unseen classes and outperforms strong discriminative baselines for few shot classification tasks. Notably, our residual pairwise network design outperforms previous state-of-the-art on the challenging mini-Imagenet dataset for one shot learning by getting over 55\% accuracy for the 5-way classification task over unseen classes.
\end{abstract}

\section{Introduction}

Human intelligence is considered the epitome towards which man-made intelligent systems strive. Two specific abilities that set the human mind apart are its ability to generalize to related but unseen data and its ability to learn from  small quantities of data. As machine learning methods have seen tremendous progress in recent years there has been growing focus on measuring the ability of models to generalize to unseen varieties of data and to learn from even very little of it. This is the aim of the one-shot learning problem \cite{lake15}. Specifically, given a dataset some of whose classes are not used for training, the model is tested for its ability to classify over those unseen classes using a single example for each class. Essentially, the model must learn features that generalize across classes.  
The one shot learning problem has been studied under various domains like Bayesian learning \cite{feifei03}\cite{salakhutdinov12} and deep representation learning using distance metrics \cite{koch15}. Deep networks till recently were felt unsuited for such a problem given the huge number of parameters involved and the possibility of overfitting on a problem with very little data. However recent work \cite{vinyals16} \cite{ravi17} has shown that deep models, especially recurrent architectures, can perform very well on the one-shot task. 

\begin{figure}
  \includegraphics[width=0.47\textwidth]{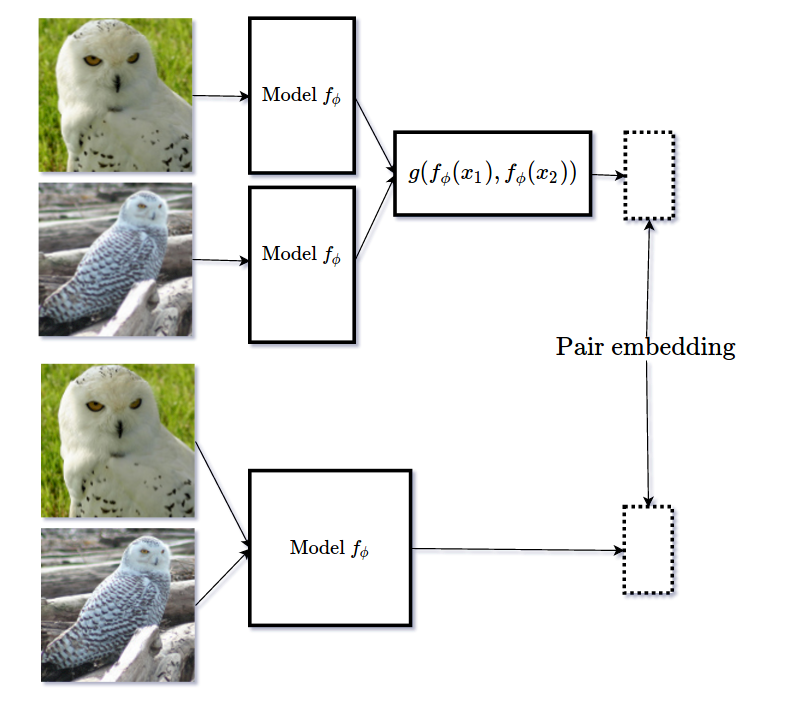}
    \caption{Two approaches to obtain the similarity embedding. In the first case $f$ can be a neural network and $g$ is a simple function like L1 or cosine distance. In the second case $g$ is also parameterized by the neural network and hence is adaptive to minimize the training objective with end-to-end optimization.}
\label{fig:1}
\end{figure}

The focus of our work, however is on two aspects which we believe are essential for a model to do well on the task of one shot learning - inferring an accurate semantic representation in a low dimensional manifold and strong regularization. Convolutional Neural Nets are very effective at both, as is shown by their success on many image classification tasks.  We additionally enforce the first criteria by presenting a hypothesis that learning an end-to-end trainable distance measure is better than a fixed distance metric (see figure \ref{fig:1}). We do this by implementing a modification of the residual network architecture \cite{he16}. Specifically our modification involves using skip residual connections and we refer to the our model as Skip Residual Siamese Network (SRPN). The model takes a pair of images and outputs a single similarity embedding vector. We train this model end-to-end for the similarity matching objective and then use it for few shot classification tasks on the Omniglot and mini-Imagenet datasets.

We propose a solution to the second problem in the form of a novel architecture derived from the Generative Adversarial Networks (GAN) \cite{goodfellow14}. The field of generative modelling has seen rapid advancement in the past few years. This can majorly be attributed to the development of various frameworks that work well in tandem with deep neural networks. Models have been developed that are able to efficiently approximate data distributions well enough to be of practical utility. An interesting subclass of such models is that of the implicit variety \cite{mohamed16} which do not prescribe a fixed parametric form for the learned distribution. These models are consequently trained using methods of comparison instead of maximizing a fixed log-likelihood function. The de facto framework for implicit models is Generative Adversarial Networks (GANs), which has been successfully used for many learning tasks. While this framework has received great interest, work towards understanding its generalization in the low-data regime and its applicability for tasks such as one shot learning has been limited. Our work attempts to bridge that gap and show that GANs provide effective regularization on unseen data distributions. Specifically, we extend the framework by trying to generate a conditional distribution which can regularize a siamese matching network. The siamese network is itself modified to incorporate a feedback that trains the generator along with the task of similarity matching. Thus, our model can be considered an instance of multi-task learning \cite{caruana98}. We show that the generated data is a strong regularizer for the similarity-matching task and helps in generalization to unseen classes.

\subsection{Related Work}
Our work is closely related to that on similarity matching using Siamese Networks \cite{bromley93}, which was extended for convolutional networks for face verification in \cite{chopra05}. Koch et. al \cite{koch15} show results using convolutional variants of these networks on the few shot classification tasks for Omniglot. Various formulations for the similarity objective such as contrastive loss \cite{hadsell06} and triplet loss \cite{weinberger09} and efficient methods to do the same \cite{schroff15} are also closely related.  

Other related approaches include those based on Deep Metric Learning such as deep embeddings for triplet loss \cite{hoffer15} and better metrics for fine-grained similarity \cite{rippel15}, and popular methods like Neighbourhood Component Analysis \cite{goldberger04}.

The few shot problem specifically has been studied from multiple perspectives, including the similarity-matching \cite{koch15}, optimization \cite{ravi17}, metric learning \cite{vinyals16}, hierarchical graphical models \cite{salakhutdinov12} etc. Our work falls in the similarity-matching paradigm ie. the model learns to classify by individually comparing and contrasting the similarities of the given class data with the test example at hand.
To the best of our knowledge, the earliest work on this problem is using Bayesian methods \cite{feifei03}\cite{feifei06}. Salakhutdinov et. al \cite{salakhutdinov12} proposed an elegant approach using hierarchical bayesian models. Lake et. al \cite{lake11}\cite{lake15} develop the framework of Bayesian Program Learning that naturally inculcates predicting by learning from small amounts of data. They also developed the Omniglot dataset used in our experiments.
Recent work includes methods that try to use deep networks for learning embeddings which are the fed to recurrent networks for calculating embeddings \cite{vinyals16}. Vinyals et. al \cite{vinyals16} also propose an approach for learning the similarity embedding based on the full context of the input set as opposed to pairwise comparisons used in typical similarity matching settings, including this work. They also modify the training procedure to be similar to the testing protocol for few shot learning. Another recent work \cite{snell17} also explicitly optimizes the one shot learning objective by essentially training the model for multiclass classification in the one shot setting. In contrast, this work aims to show that deep residual networks are naturally well suited to generalization objectives and even the proper pairwise similarity matching objective can perform the one shot task well. The recent work in \cite{shyam17} proposes a recurrent model that also outputs a pairwise embedding, however their reasoning is towards using attention to selectively infer the next step whereas we are focused on a natural, scalable extension to the CNN for one-shot learning. 

Another related area is that of deep generative models for one-shot learning. Edwards et. al \cite{edwards16} propose handling the one shot classification task by learning dataset statistics using the amortized inference of a variational auto-encoder. Another popular framework of Generative Modeling - GANs \cite{goodfellow14} and their conditional \cite{mirza14} \cite{gauthier14} alternatives are the essential components of our Generative Regularizer. Our design of using the generator with a discriminator is inspired by the ImprovedGAN model \cite{salimans16} for semi-supervised learning. Another work which uses generated images for improving one shot learning is \cite{hariharan16}, although they train their model as a transformer for test time generations using cosine similarity in the training set to generate analogies instead of an adversarial loss.\\

\subsection{Contributions}
To summarize, our work proposes a model for similarity matching with two suggested improvements for similarity measures and . The specific contributions are: 
\begin{itemize}
    \item We propose using a trainable distance measure for the task of one shot learning, and our implementation based on the modification of the residual network achieves state-of-the-art on the challenging mini-Imagenet dataset
    \item We show that generated data acts as a strong regularizer for the task of similarity matching and design a novel network based on the GAN framework that shows improvements for the one shot learning task

\end{itemize}
\section{Preliminaries}

\subsection{Few Shot Learning}
The objective of few shot learning is to measure the ability of a model to learn generalizable features across classes unseen during training. Specifically, the model is trained on labelled training data $(x,y)$ where each example comes from a subset of the total classes $C_{train} \subset C$. Then during the testing phase, the model is provided with a single example from each of the chosen classes in $C_{test} \subset C - C_{train}$ from the disjoint testing set. Using these examples only as the support set $S$, the model must correctly classify another sample chosen from $C_{test}$. The number of classes in $C_{test}$ (generally 5 or 20) and the number of examples (generally 1 or 5) characterize the problem.

\subsection{Generative Adversarial Networks}
Generative Adversarial Networks (GANs) propose a method for learning a continuous-valued generative model without a fixed parametric form for the output. This is done by establishing a generator function that maps from latent to data space and a discriminator function that maps from the data space to a scalar. The discriminator function $D$ tries to predict the probability of the input being from $p_{data}$, and the generator is trained to maximize the error of the discriminator on $G(z)$. Both the generator and the discriminator are parametrized as deep neural networks. The value function of this min-max formulation can be written as follows: 

\begin{multline}
    \text{min}_\theta \text{max}_\phi\mathcal{V} = \mathbb{E}_{x \sim p_{data}}\ \mathrm{log} (D_{\phi} (x)) \\ + \mathbb{E}_{x \sim G_{\theta}(z)}\ \mathrm{log} (1 - (D_{\phi} (x)))
\end{multline}

\begin{figure*}
  \includegraphics[width=\textwidth, height=5.5cm]{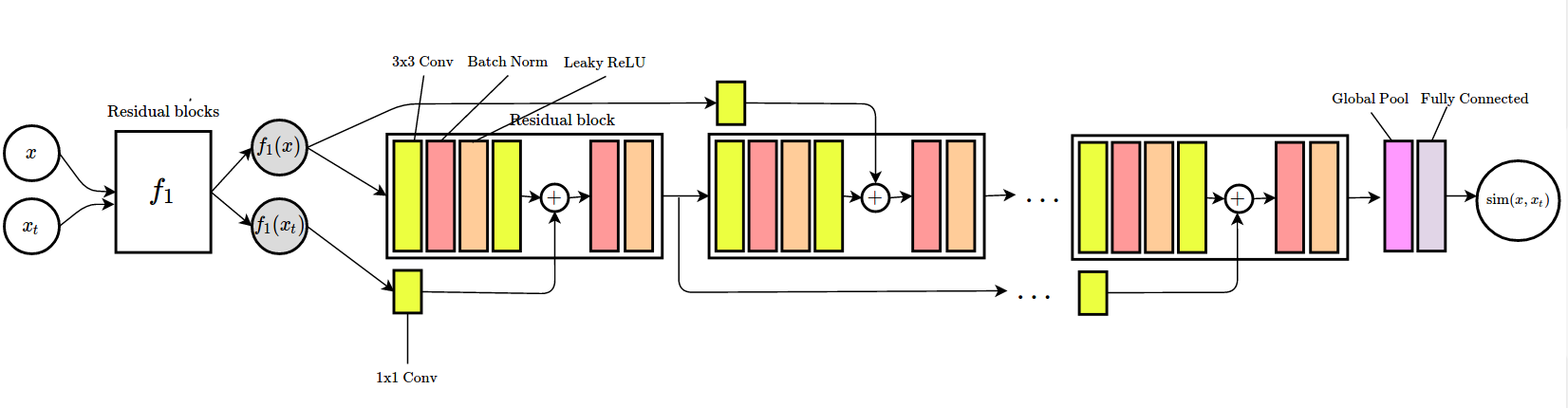}
    \caption{Our proposed model Skip Residual Pairwise Net (SRPN). The network separates the intermediate computations for the inputs $x$ and the image being compared $x_t$, which are then passed through separate pathways using the residual connections. The final output is a single similarity vector for the pair where the distance measure is itself learned by the network.}
    \label{fig:2}

\end{figure*}

\section{Models}
\subsection{Skip Residual Pairwise Network}
Siamese networks in literature have been typically referred to models that have two identical feed-forward paths to generate embeddings for two data-points and the similarity is then computed as a function $g$ of these two embeddings. While this is a successful paradigm, one can argue that the choice of $g$ is arbitrary and the model is forced to learn an embedding that works well with a particular choice of $g$. Such a choice may often be sub-optimal, such as when the embedding itself is an intermediate step in a solving a problem such as similarity matching for one-shot learning. To circumvent this we propose a pairwise network which is takes as input a pair of data points and is trained end-to-end to optimize a particular objective - in our case the similarity between two data points. Note that this leads to the loss of symmetric property necessary for metrics. \\
Our network design is presented in figure \ref{fig:2}. The initial part of the network focusing on most abstract features is similar to an ordinary residual network. Then, the intermediate embeddings are split for the input pair and are fed through different pathways alternating between a residual skip connection using $1\times1$ convolutions and two $3\times3$ convolutions with Batch Normalization \cite{ioffe15} and a non-linearity. The network is designed thus to ensure 2 objectives:
\begin{itemize}
\item ensure adequate mixing ie. allow the intermediate representation of one data point to affect the intermediate representation of the other data point, instead of mixing only at the final layer as in normal siamese networks 
\item maintain the residual structure which has been successfully trained for very deep networks
\end{itemize}

The final output of our model is then a single embedding which is fed into a linear classifier to predict the similarity. The whole network is trained end-to-end using the binary cross-entropy loss similar to a normal siamese network.\\

\textbf{Multiplicative Units:} Multiplicative gating units have been shown to be effective in learning with deep models \cite{srivastava15}. Multiplicative interactions could improve performance in our case by allowing better mixing of information from the two sources. We followed the formulation used in \cite{kalchbrenner16} to replace the second rectified linear unit activation in our block with a three-gated multiplicative unit. However, we found it did have any significant positive effect on the mini-Imagenet experiments while increasing the number of parameters substantially, thus we refrain from using it in our final model.

\subsection{Generative Regularizer}
\begin{figure*}
  \includegraphics[width=0.9\textwidth]{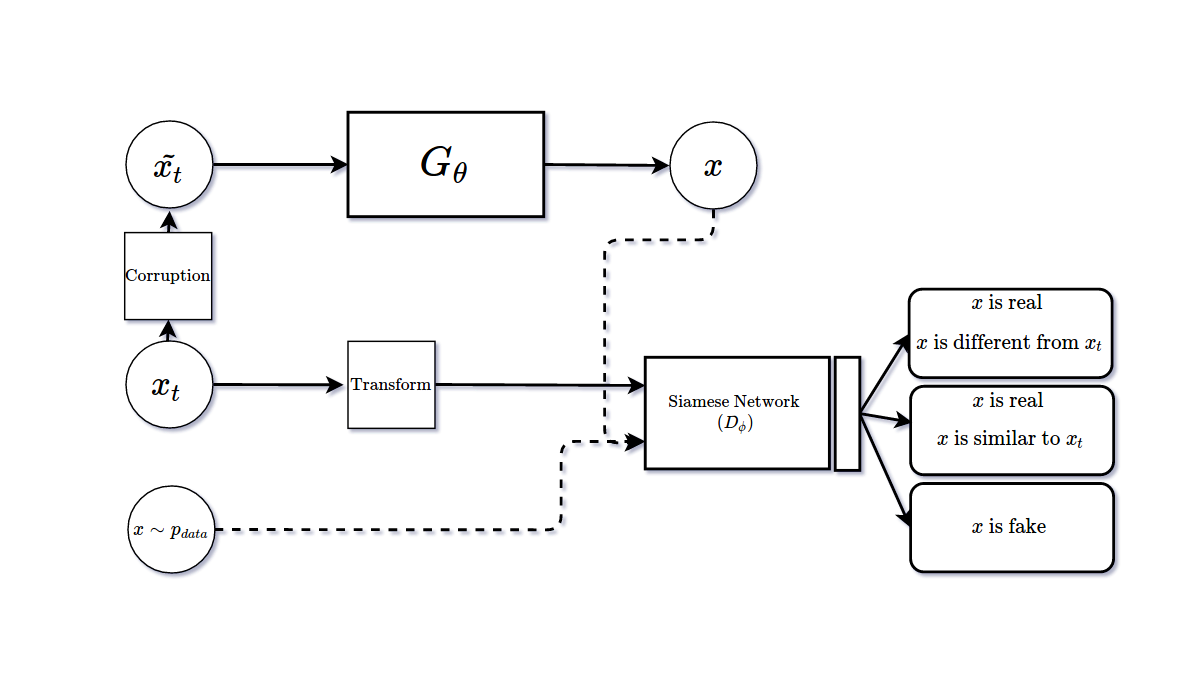}
    \caption{An indicative figure of the proposed Generative Regularizer model. The Discriminator $D\phi$ takes two samples $x,x_t$ and outputs a probability distribution over the 3 indicated output classes. The conditioning sample $x_t$ is spatially transformed and fed to the discriminator. The other data point $x$ comes from either the Generator which takes as input the corrupted conditioning data point $x_t$ or from real data samples. Since the labels are known, the real data point $x$ is chosen with equal probability from the same class as $x_t$ or a different one.}
    \label{fig:3}
\end{figure*}

We adapt the Generative Adversarial Network to create the second model whose purpose is to provide better regularization for the similarity matching task. The proposed model consists of two networks - a similarity matching discriminator network $D_\phi(.)$ and a generator network $G_\theta(.)$ where ${\phi,\theta}$ are the parameters of the networks. The objective of the discriminator network is to predict whether an input data point $x$ belongs to the same class-conditional distribution as another data point $x_t$. The discriminator is simultaneously trained to classify the generated images as fake. The generator is trained with the objective of being able to generate plausible variations of a given image - ie. images that belong to the same class conditional distribution as the input image $x_t$. The structure of our model is depicted in figure \ref{fig:3}. The mathematical formulation of the two networks is described below.

\textbf{Discriminator Network:} The discriminator network $D_\phi(x,x_t)$ tries to predict the probability whether the input data point $x$ belongs to the same class-conditional distribution as $x_t$, ie. $x \sim p_{data}(x|class(x_t))$ where $class(x_t)$ is the class of the data point $x_t$. Thus, the network tries to maximize the probability for $x \sim p_{data}(x|class(x_t))$ and minimize it for $x \sim p_{data}(x|class \neq class(x_t))$ and $ x \sim p_{gen}(x|G(\tilde{x_t}))$.  We avoid using explicit label vectors for $class(x_t)$ chiefly because of 2 reasons: i) we do not want the network to memorize class details but instead only the invariant transformations ii) it would prevent conditioning the generator over classes unseen during training. The network is simultaneously trained to minimize  with $x$ drawn from any other distribution. It thus also acts as the similarity-based classifier where similarity is ascertained from class labels. \\
For our purpose we found that having the discriminator separate outputs for $p_{data}(x|class \neq class(x_t))$ and $p_{\theta}(x|\tilde{x_t})$ leads to better performance. This is consistent with the work of \cite{salimans16} for semi-supervised learning. Hence we formulate our discriminator to minimize the following loss: 
\begin{multline}
    \mathcal{L}_{dis} = - \mathbb{E}_{(x,x_t,y) \sim p_{data}}\ \mathrm{log} (p_{\phi} (y|x,x_t)) \\ - \mathbb{E}_{x \sim p_{\theta}}\ \mathrm{log} (1 - p_{\phi} (y=0,1|x,x_t))
\end{multline}
where $y=1$ is the label associated with the tuple $(x,x_t),\ x \sim p_{data}(x|\mathrm{class}(x_t))$  and $y=0$ is associated with $x \sim p_{data}(x|\mathrm{class} \neq \mathrm{class}(x_t))$. \\

\begin{figure}
  \includegraphics[width=0.47\textwidth]{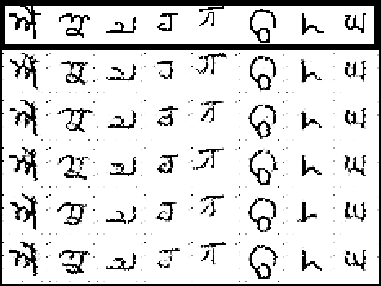}
    \caption{One-shot generation results over unseen classes. The first row is of the samples from the test set used to condition the generator.}

\end{figure}
\textbf{Generator Network:} The generator in our proposed model is a parametric function $G_\theta(.)$ that maps from the data space to another point in the same space, ie. it acts as an auto-encoder. To prevent the generator from simply copying the input we enforce regularization by corrupting the input data point $x_t$ by some stochastic process. The corrupted input data point $\tilde{x_t}$ is then down-scaled by passing through a series of convolutions followed by up-scaling using a series of transposed convolutions. Essentially our generator behaves as a denoising autoencoder \cite{vincent10} but instead of a reconstruction loss it is trained with an adversarial loss. The generator is trained to minimize the following objective:  
\begin{multline}
    \mathcal{L}_{gen} = - \mathbb{E}_{x \sim p_{data}, \tilde{x} \sim \mathcal{C}(\tilde{x}|x)}\ \mathrm{log} (p_{\phi} (y=1|G_\theta(\tilde{x}), x_t))
\end{multline}

\section{Experiments}
We perform experiments to validate the proposed models for few shot learning over two datasets - Omniglot and mini-Imagenet. All our models are implemented using the Lasagne library \cite{lasagne} for Theano \cite{theano}.

\subsection{Omniglot}
Omniglot was introduced in \cite{lake15} for the express purpose of measuring one shot learning performance of models. It consists of 1623 classes of characters with 20 binary images per class. Results on the dataset have been reported using two different configurations in literature - a within alphabet setting of \cite{lake15} and the more recent across alphabet setting of \cite{vinyals16}. In this paper we follow \cite{vinyals16} to have consistent results with other recent works. \\
The dataset is divided into two parts - the first 1200 classes are used for training and validation, while the remaining are test classes for the few shot tasks. For this dataset we follow the training protocol (algorithm \ref{algo:1}) with $\text{num tests} = 200$ and $\text{runs per test} = 20$. \\
We tested both our proposed improvements, Skip Residual Pairwise Networks (SRPN) and Generative Regularizer (GR) on this dataset. 

We train two architectures for the convolutional siamese network: i) $\text{Siam-I}$: a smaller one with 5 convolutional layers and a final global pooling layer ii) $\text{Siam-II}$: a larger network with 5 residual blocks followed by global pooling. We follow the Wide Residual Networks design with k=2 \cite{zagoruyko16}. We modify $\text{Siam-II}$ for the design of our SRPN model. While the depth of $\text{Siam-I}$ and $\text{Siam-II}$ is different, both have a similar number of trainable parameters. We also use these convolutional siamese networks as baselines for our proposed models.  

To ensure consistency with other published results, we rescale the images to $28$x$28$ pixels, and augment the training data with random rotations (by $\pm 45$ degrees and/or translations ($6$ pixels) in both X and Y axis. The training is done using mini-batch gradient descent with Adam \cite{kingma14} updates and batch size of $128$. The initial learning rate is set to $8\times10^{-4}$ and $\beta_1=0.5$ for GR experiments. $\text{L2}$ regularization is used with all models except those with GR. We maintain a validation set of $60$ classes from training for early stopping. Results are reported in table \ref{table:1}

\begin{algorithm}
\SetAlgoLined
total-correct-pred = 0\;
 \For{num-tests}{
  sample support set $S$ by randomly picking $N$ classes and $1$ example for each of those classes\;
   \For{runs-per-test}{
    sample a test case $x_t$ from one of the classes in $S$\;
    \If{predict($S$,$x_t$) == class($x_t$)}
    {    total-correct-pred += $1$
    }
  }
 }
 \Return{(total-correct-pred / (num-tests*runs-per-test))}\\
 \vspace*{5mm}
 \caption{Testing protocol for $N$-way one shot learning}
 \label{algo:1}
 
\end{algorithm}

\subsection{Mini-Imagenet}
Mini-Imagenet was introduced recently in \cite{vinyals16} as a more challenging dataset for one shot learning tasks. The dataset consists of 100 classes of natural images from the Imagenet dataset \cite{russakovsky15} with 600 RGB images per class, rescaled to $84$x$84$ pixels. Since the standard splits for the dataset were not released, researchers have reported results by randomly selecting 100 classes from the Imagenet dataset \cite{ravi17}\cite{snell17}. We follow a similar practice here. \\

\begin{figure}
  \includegraphics[width=0.47\textwidth]{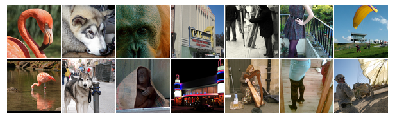}
    \caption{Actual data samples from some of the test classes for the Mini-Imagenet Dataset. Each column shows two images from the same class. Given the huge inter-class variation in these natural images, the problem of recognizing the class by comparing it to a single image of each of the other classes is challenging.}

\end{figure}

The dataset is divided into two parts - the first 80 classes are used for training and validation, while the remaining are test classes for the few shot tasks. For this dataset we follow the training protocol (algorithm \ref{algo:1}) with $\text{num-tests} = 100$ and $\text{runs-per-test} = 100$. \\
We tested both our proposed improvements, Skip Residual Pairwise Networks (SRPN) and Generative Regularizer (GR) separately on this dataset. However we found that Generative Regularization (GR) did not yield any benefits for these very deep models as the generator itself was not able to learn well. We believe it is due to the unstable nature of GAN training. Hence we only report results on Skip Residual Pairwise Networks (SRPN) for this task.

Since network depth has been shown to be essential for good performance on Imagenet, we restrict ourselves to the Wide Resnet model \cite{zagoruyko16} with network depth of 40 and k=2. We train a siamese network using this model as our baseline and we modify it for the design of our SRPN model to ensure the results are comparable. 

To ensure consistency with other published results, we rescale the images to $84$x$84$ pixels. No data augmentation or pre-processing of any kind is done. The training is done using mini-batch gradient descent with Adam \cite{kingma14} updates and batch size of $64$. Both models are trained for $100000$ updates. The initial learning rate is set to $5\times10^{-4}$, and is annealed linearly to $1\times10^{-4}$. $\text{L2}$ regularization is used for both models with initial value of $5\times10^{-7}$ increased to $1\times{10}^{-6}$ after $60000$ updates. Results are reported in table \ref{table:2}.

\begin{table}[h!]
\begin{tabular}{ |p{5cm}||p{1cm}| } 
 \hline
 Model & 1 shot \\ [0.5ex] 
 \hline\hline
 Pixel Distance \cite{vinyals16} & 26.7 \\ 
 Matching Nets w/ Finetuning \cite{vinyals16} & 93.5 \\ 
 Neural Statistician \cite{edwards16} & 93.1 \\
 Conv ARC \cite{shyam17} & \textbf{97.5} \\
 Prototypical Networks \cite{snell17} & 96.0 \\ 

 \hline
 Baseline (Siam-I) & 88.4  \\ 
 Baseline (Siam-II) & 92.0 \\
 \textbf{GR + Siam-I} & 93.6 \\
 \textbf{GR + Siam-II} & 91.2 \\
 \textbf{SRPN} & \textbf{94.8} \\ 
 \hline
\end{tabular}
\vspace*{5mm}
\caption{Accuracy (\%) for the 20-way one shot learning experiment on the Omniglot dataset.}
\label{table:1}
\end{table}

\begin{table}[h!]
\begin{tabular}{ |c||c|c| } 
 \hline
 Model & 1 shot & 5 shot \\ [0.5ex] 
 \hline\hline
 Pixel Distance \cite{vinyals16} & 23.0 & 26.0 \\ 
 Baseline w/ Nearest Neighbour \cite{ravi17} & 41.1 & 51.0  \\
 Matching Nets FCE \cite{vinyals16} & 46.6 & 60.0 \\ 
 Meta-Learner LSTM \cite{ravi17} & 43.4 & 60.6 \\
 Conv ARC w/ L2 Reg \cite{shyam17} & 49.1 &  -   \\
 Prototypical Networks \cite{snell17} & 49.4 & 68.2 \\ 

 \hline
 Baseline (Wide Resnet depth=40) & 50.7 & 66.0 \\ 
 \textbf{SRPN} & \textbf{55.2} & \textbf{69.6} \\ 
 \hline
\end{tabular}

\vspace*{5mm}
\caption{Accuracy (\%) for the 5-way few shot learning experiments on the mini-Imagenet dataset. Note that all models have reported results using a random 100 class subset of the Imagenet dataset.}
\label{table:2}
\end{table}

\section{Discussion}
\begin{figure}
  \includegraphics[width=0.47\textwidth]{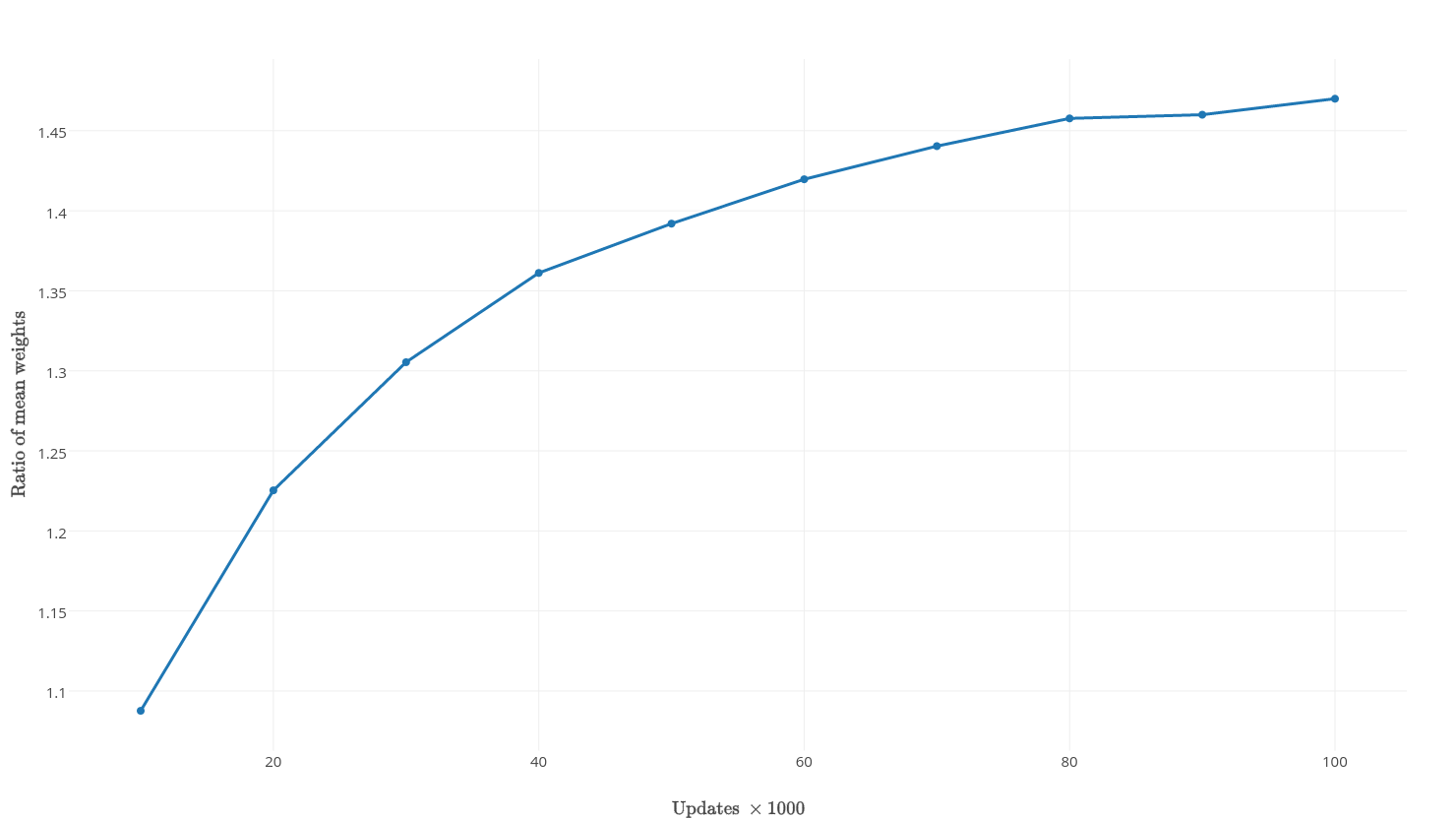}
  \caption{Ratio of mean weights of the Wide Residual networks and the modified SRPN as training progresses on mini-Imagenet}\label{fig:plot1}

\end{figure}

\begin{figure}
  \includegraphics[width=0.47\textwidth]{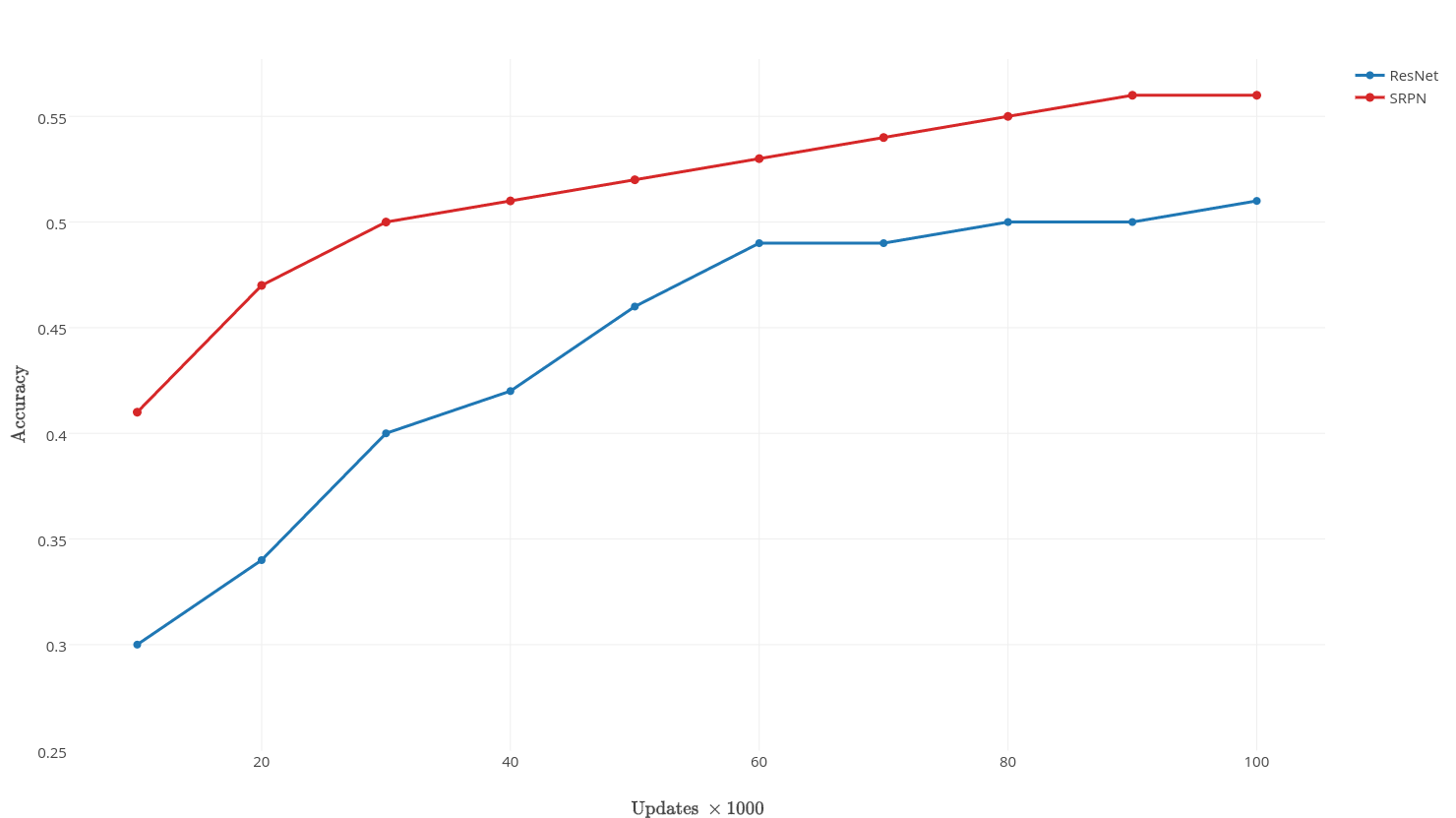}
  \caption{One shot test accuracy comparison for the SRPN (red) and the wide residual siamese net (blue) as training progresses on mini-Imagenet}\label{fig:plot2}

\end{figure}

\begin{figure}
  \includegraphics[width=0.47\textwidth]{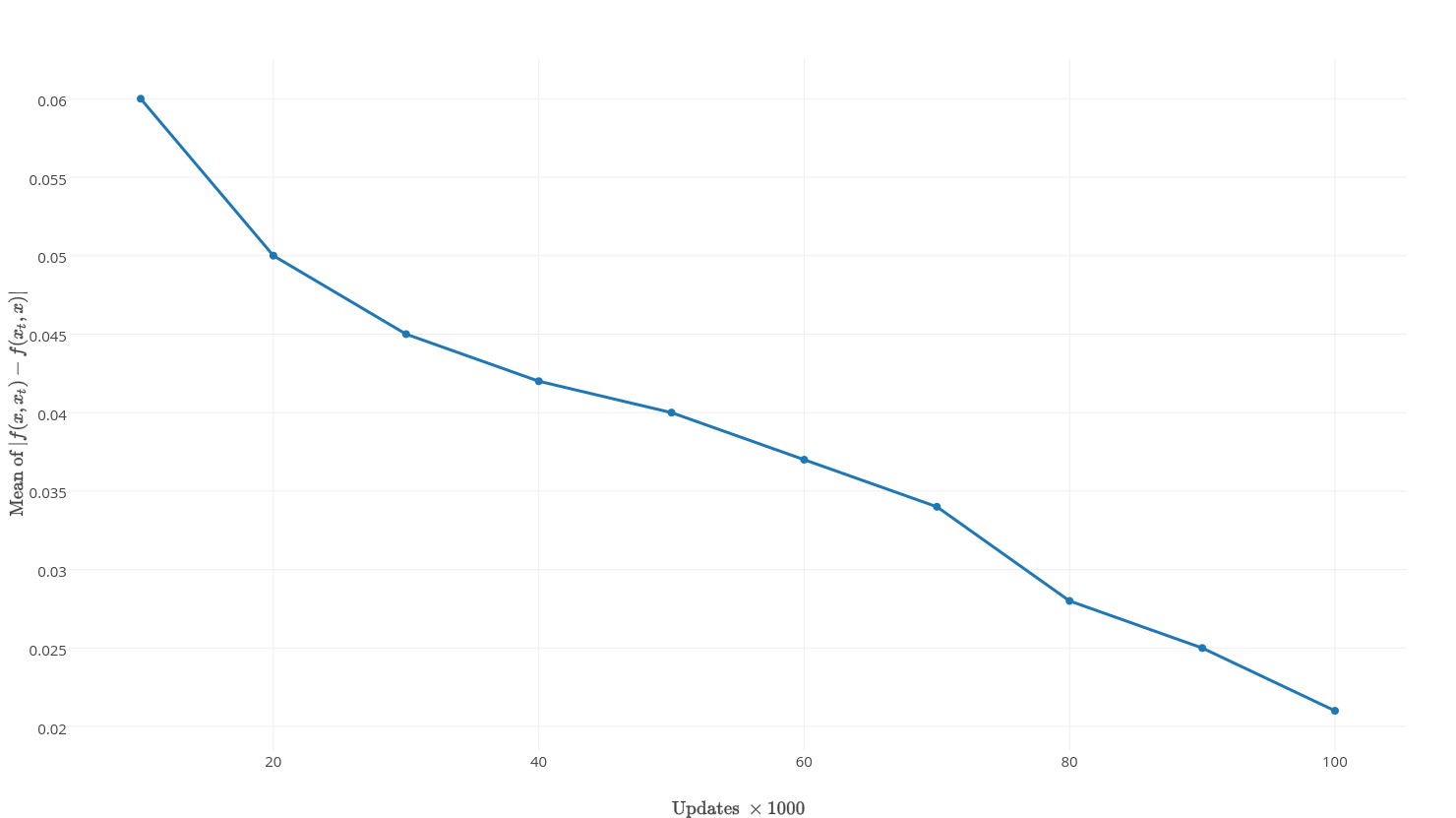}
    \caption{Mean of the difference between embeddings for symmetric inputs for the SRPN model on mini-Imagenet ie. $|f(x,x_t) - f(x_t,x)|$ as training progresses, on the mini-Imagenet dataset.} \label{fig:plot3}

\end{figure}

We discuss some observations about our two suggested models - the Skip Residual Pairwise Network (SRPN) and the Generative Regularizer (GR). First, we notice that not only does the SRPN achieves higher accuracy than the equivalent Residual Siamese Network but its convergence is much faster and the mean network weight is noticeably smaller. (see figure \ref{fig:plot1},\ref{fig:plot2}) In our opinion, this happens because the SRPN is not forced to learn an embedding that works well with a fixed distance metric, instead it is able to adapt to the distance measure that best minimizes the total loss. This allows the network to find a manifold which reduces the similarity loss as well as the regularization penalty, effectively leading to better generalization performance. 
We also note that while the SRPN is not explicitly trained to learn a symmetric embedding the model does this automatically as can be seen by the diminishing difference between the embeddings (pre-final layer) in figure \ref{fig:plot3}.
Also interesting to note is that for the mini-Imagenet task our siamese residual net baseline outperforms the previous state-of-the-art, which reinforces the importance of depth in complicated image recognition tasks but is also an indicator that deep convolutional models with millions of parameters can be successful in learning from features that generalize well to unseen data distributions and thus do well in the one shot learning setup.  

Our experiments with the Generative Regularizer (GR) give a mixed-bag of results - the regularization is much more effective than $L2$ regularization as shown by the results on Omniglot with $\text{Siam-I}$ model reaffirming the results in semi-supervised learning with such methods \cite{salimans16}. However, training the generator-discriminator framework is challenging due to its sensitivity to hyperparameters, often leading to poorly trained generator that impedes training as seen with very deep models as discriminators. We believe that advances in the GAN training methodology could be applied here directly and that a properly trained GR+SRPN network would outperform one with L2-regularization. 

As an aside, we also tried training the GR model without providing conditioning information, ie. mapping from Gaussian noise to the image space. We observed that the generator inevitably collapsed to a single point, but was still able to regularize the siamese network. This indicates that there are multiple failure modes of the generator and that good generation is not essential for regularization. We will continue to experiment in this direction. 

\section{Conclusion}

To summarize, we identified fixed distance measures and weak regularization as major challenges to similarity matching and its extension to one shot learning and presented a network design for each of the problems. Our Skip Residual Pairwise Network outperforms an equivalent Residual Siamese Network and achieves state of the art performance on the mini-Imagenet one shot classification dataset. Our Generative Regularizer shows promising results and outperforms $L2$-regularization on the Omniglot dataset. Future work would focus on integrating the two networks presented here by tapping into advancements in the training of Generative Adversarial models.

{\small
\bibliographystyle{ieee}
\bibliography{egbib}
}

\end{document}